\documentclass[
twocolumn, 
hf, 
]{ceurart}
\usepackage{xcolor}

\usepackage{listings}
\begin{document}

\copyrightyear{2025}
\copyrightclause{Copyright for this paper by its authors.
  Use permitted under Creative Commons License Attribution 4.0
  International (CC BY 4.0).}

\conference{Preprint}

\title{Can Large Language Models (LLMs) Describe Pictures Like Children? A Comparative Corpus Study}


\author[1]{Hanna Woloszyn}[%
orcid=0009-0007-6641-4790,
email=hwoloszy@uni-koeln.de
]
\cormark[1] 
\address[1]{Self-Learning Systems Lab, Faculty of Human Sciences, Department of Special Education and Rehabilitation, Cologne, Germany}

\author[1]{Benjamin Gagl}[%
orcid=0000-0002-2339-6293,
email=benjamin.gagl@uni-koeln.de,
]
\cormark[1] 



\cortext[1]{Corresponding author.}

\begin{abstract}
The role of large language models (LLMs) in education is increasing, yet little attention has been paid to whether LLM-generated text resembles child language. This study evaluates how LLMs replicate child-like language by comparing LLM-generated texts to a collection of German children's descriptions of picture stories. We generated two LLM-based corpora using the same picture stories and two prompt types: zero-shot and few-shot prompts specifying a general age from the children corpus. We conducted a comparative analysis across psycholinguistic text properties, including word frequency, lexical richness, sentence and word length, part-of-speech tags, and semantic similarity with word embeddings. The results show that LLM-generated texts are longer but less lexically rich, rely more on high-frequency words, and under-represent nouns. Semantic vector space analysis revealed low similarity, highlighting differences between the two corpora on the level of corpus semantics. Few-shot prompt increased similarities between children and LLM text to a minor extent, but still failed to replicate lexical and semantic patterns. The findings contribute to our understanding of how LLMs approximate child language through multimodal prompting (text + image) and give insights into their use in psycholinguistic research and education while raising important questions about the appropriateness of LLM-generated language in child-directed educational tools.
\end{abstract}


\maketitle

\section{Introduction}

Large language models (LLMs) have impacted fields such as corpus linguistics, psycholinguistics, and natural language processing (NLP) by providing new ways to generate text in response to prompts \citep[e.g.,][]{brown2020language,schepens2023can}. LLMs' accessibility and user-friendliness have supported their growing use in research and applications, such as chatbots \citep[e.g.,][]{brown2020language}. Recent work has explored using LLMs to support children's creative writing skills \citep{elgarf2024fostering}, suggesting that LLM-generated output may scaffold and enhance children's writing. However, these findings also highlight that LLM-generated texts lack lexical richness, raising questions about their effectiveness as linguistic models in educational contexts \citep{liu2024benchmarking, schepens2023can}. Here, we extend this evidence based on the state-of-the-art models that offer image prompt capabilities, generating text from text and visual prompts \citep{tsimpoukelli2021multimodal,alayrac2022flamingo}. These LLMs allow us to generate a unique yet child-specific corpus of texts that simulate children's responses to a set of picture stories with accompanying text prompts \citep{laarmann2019litkey}. 

The study of child language and first language acquisition is foundational in linguistics, providing insights into cognitive development and mechanisms of language learning \citep[e.g.,][]{tomasello2005constructing}. Children undergo significant linguistic and interactive development during early childhood, making this period critical for understanding human language acquisition. Although, among others, classic German child-language corpora have supported acquisition research \citep{schroeder2015childlex, macwhinney2000childes}, their use in NLP remains limited due to ethical, practical, and availability constraints \citep{casillas2017new}. LLMs primarily rely on large-scale datasets dominated by adult and high-resource language data \citep{luo2023perspectival}. Child language remains marginal in the models’ training data, especially in underrepresented languages (but see, e.g., \citealp{warstadt2023findings}). As a result, there is limited research on whether LLMs can efficiently generate child-like text that mimics the linguistic patterns and conceptual structures of children’s writing (but see \citealp{liu2024benchmarking,schepens2023can}).

Child language acquisition is a multimodal process \citep{morgenstern2023children, karadoller2024first,vigliocco2014language}. From relating spoken words to objects in infancy, to multimedia aids helping children learn in school, the co-occurrence of visual and auditory language-related stimulation is critical for effective learning (e.g., \citealp{andra2020learning}). One particularly intriguing case is writing tasks. Here, children either describe a visual stimulus or should be able to write creative texts in response to a visual image after the instruction of, for example, a teacher \citep{berkling-2016-corpus,berkling20182nd,becker2022scriptoria}. In our study, we exploit the recently developed multimodal capabilities of an LLM that allow text and visual stimulation. Using multimodal prompting (text and image), we can investigate if LLMs can generate child language based on complex visual stimulation, by comparing the texts to texts from children in response to the same images. Such approaches offer insights into how LLMs could support child language development, creative writing based on modalities different than text, as such modalities are common in children's learning, yet still used much less in current LLM research due to its predominant focus on textual input.

To address this gap, we compare LLM-based corpora to the Litkey Corpus of German children’s texts \citep{laarmann2019litkey}. This comparison offers insights into how well LLMs simulate child language and provides a framework for understanding the implications of using LLM-generated corpora in linguistic and NLP contexts. Moreover, in psycholinguistic research, corpora generated by LLMs for underrepresented languages and groups have the potential to expand current research beyond the predominantly studied populations and high-resource languages \citep[e.g.,][]{gagl2022eye}.

Our study investigates whether LLMs and children generate similar descriptions of the exact same picture stories used in the Litkey Corpus (stories can be downloaded from: \href{https://www.linguistics.rub.de/litkeycorpus/access.html}{Litkey Corpus}). A comparative analysis examines the corpora across three linguistic levels: word-level (word frequency \citep{schepens2023can,schroeder2015childlex,gregorova2023access}, lexical diversity \citep{schepens2023can} and word-length in letters), syntax-level (i.e., parts-of-speech; POS \citep{martinez2012part} and sentence-length), and semantic-level characteristics (i.e., vector-based semantic similarity with word embeddings \citealp{gunther2019vector}). The study investigates whether LLMs can replicate child language and assesses potential risks for educational use. By analyzing these patterns, we expect to uncover both similarities and key differences that provide insight into the capabilities and limits of LLMs in modeling child language.

\section{Methods}

\begin{figure}[tb]
  \centering
  \includegraphics[width=.6\linewidth]{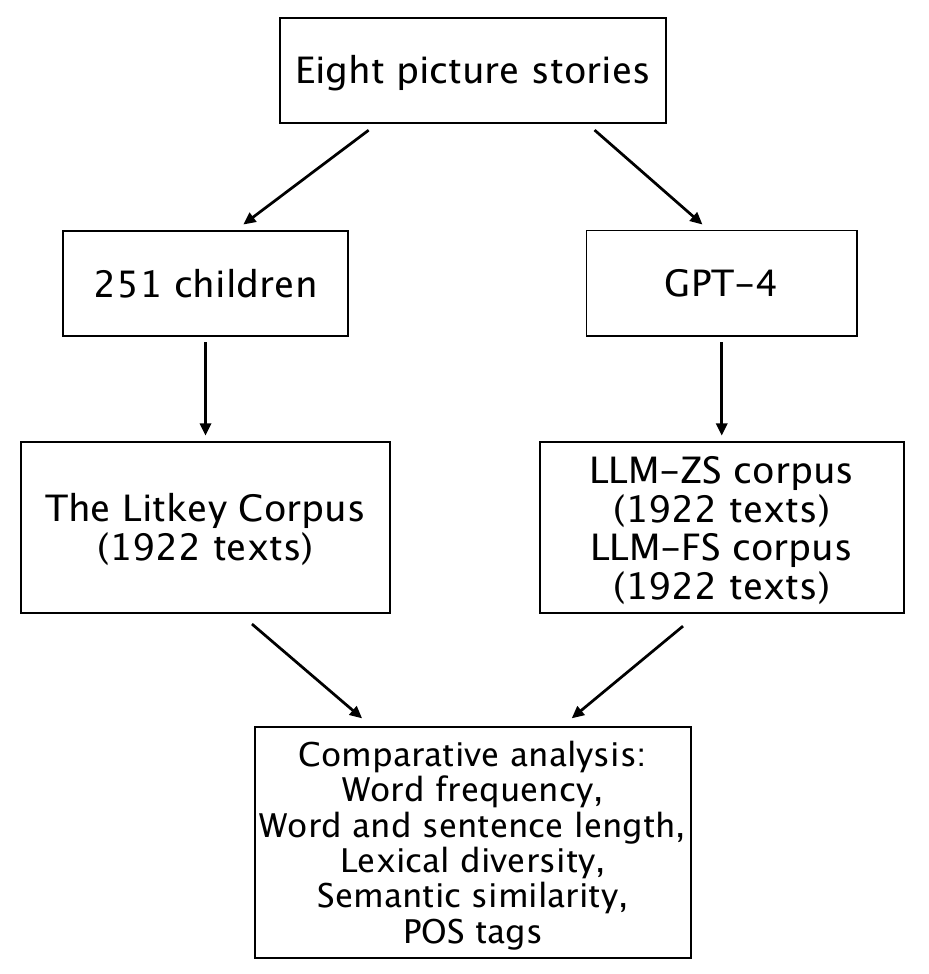}
  \caption{Methods overview: Comparison of the Litkey Corpus with two LLM-based corpora based on zero-shot prompting (LLM-ZS) and few-shot prompting (LLM-FS) method generated from the same picture stories.}
  \label{fig:methods_structure}
\end{figure}

Figure~\ref{fig:methods_structure} provides an overview of the current study. Using identical visual stimuli (eight picture stories; see the \href{https://www.linguistics.rub.de/litkeycorpus/access.html}{Litkey Corpus}), written texts were produced by 251 children (i.e., the Litkey Corpus; \citealp{laarmann2019litkey}) and by an LLM (GPT-4; \citealp{achiam2023gpt}) using two different prompt designs. The first prompt presented each picture story individually and the mean age of participating children; this formed the LLM zero-shot corpus (LLM-ZS). The second prompt also included the mean age but also two randomly drawn example pairs of picture stories and child descriptions from the Litkey to create few-shot corpus (LLM-FS). The goal here was that the model better adapts to child language. In the LLM-FS corpus, the prompt specified that the children's output modality was writing. After corpus generation, we systematically compared the corpora based on well-established psycholinguistic measures on the word-, syntax- and semantic level. Word-level characteristics were: word frequency, word length, and lexical diversity. Syntax-level characteristics were sentence length and parts-of-speech tags. We estimated semantic similarity based on vector space representations from word embeddings to compare semantic text characteristics. All resources, including source code and generated corpora, are publicly available at: \href{https://osf.io/ntb25/}{OSF Repository}.

\subsection{The Litkey Corpus}
The Litkey Corpus is a richly annotated collection of German children’s texts designed to highlight the later stages of orthographic and literacy development \citep{laarmann2019litkey}. The corpus enables a comprehensive analysis of language acquisition across diverse learner backgrounds \citep{laarmann2019making}.

The corpus consists of texts written by 251 German primary school students, with an average age of the participants of 9.6 years (grades 2–4). It is based on eight picture stories comprising six text-less images (i.e., each picture story presents all six images at once) featuring three recurring protagonists (a boy, Lars, a girl, Lea, and a dog, Dodo). Although the characters are recurring, the stories are independent of each other. The corpus consists of 1,922 texts with 212,505 tokens and 6,364 types. The collection process involved experimenters or teachers instructing the task, including naming the stories’ main characters, followed by a 10-minute discussion of the picture stories among the children without adult involvement. The children then described the stories in a 30-minute writing period. Thus, we assume the Litkey Corpus is a robust resource for our planned comparison study.

\subsection{Large Language Model}
For the generation of the LLM-ZS corpus, we used GPT-4V, a transformer model that supports text and image prompts \citep{achiam2023gpt}. As the LLM-FS corpus was generated later (April 2025), the GPT-4V model was deprecated (i.e., see \href{https://platform.openai.com/docs/deprecations}{OpenAI deprecation page}). Instead, we used GPT-4o model, a recommended replacement by OpenAI for the previous model. A common problem with high-end LLMs is that key architectural details and model weights are not disclosed; however, at the time (March 2024), no preferred open-source alternatives were available. Models were accessed via the GPT API in Python.

\subsection{Prompt Engineering and the LLM-based Corpora}
Prompt engineering is beneficial for generating text that replicates child-specific language (e.g., see  \citealp{schepens2023can, liu2024benchmarking}). Here, we prompt the LLMs to generate child-like descriptions of the Litkey picture stories in German. In the prompt, we apply multimodality through the use of both image and textual input. Each prompt includes a Base64-encoded image from the picture stories included in the Litkey Corpus, together with a text-based instruction to simulate children’s picture descriptions. The model input combines both the visual image and the instructional text.

For the zero-shot corpus (LLM-ZS), we included the children’s average age (9.6 years) to better align outputs with the original data. We also provided a prompt with paired examples of image-description for the few-shot corpus (LLM-FS). After testing different prompt structures to ensure the model generated language replicating written descriptions of children, rather than simulating or “imagining” how a child might write, we developed two prompts that combine visual input (i.e., picture stories) encoded in Base64 format. 
Below is the prompt used for the LLM-ZS corpus: 

\vspace{0.6em}
\noindent
\texttt{“Du bist ein 9.6-jähriges Kind.} \texttt{Wie würdest du dieses Bild beschreiben?”} (“You are a 9.6-year-old child. How would you describe this picture?”).
\vspace{0.6em}
\noindent



For the few-shot corpus, we included two randomly drawn example image–description pairs from the Litkey in each prompt. To avoid reducing the linguistic variability in the model's output, the examples were selected from picture stories other than the target story. For example, if the target story to be described by the model was \textit{Eis}, the examples would include any other picture story and its corresponding description, except for \textit{Eis} itself. This prevented the model from imitating language tied to a specific story. 
The prompt was phrased as follows:

\vspace{0.6em}
\noindent
\texttt{“Du bist ein 9.6-jähriges Kind. Hier sind einige Bildbeschreibungen von anderen Kindern zu anderen Bildergeschichten:”}, \texttt{“Wie würdest du dieses Bild beschreiben?”} (“You are a 9.6-year-old child. Here are some picture descriptions from other children about different picture stories.”, “How would you describe this~picture?”)
\vspace{0.6em}
\noindent

For the prompts, we adjusted the following parameters: the \texttt{max\_tokens} parameter was set to 2,000 to allow for sufficient text length, while the \texttt{temperature} was set to 0.7 to encourage varied but coherent outputs (see  \citealp{schepens2023can} for a indication that setting the parameter to that value could be beneficial for an adequate text generation). We increased the \texttt{max\_tokens} parameter to 5,000 for the few-shot corpus, as the examples included in the prompt were counted into the token limit. All other parameters (e.g., \texttt{presence\_penalty}, \texttt{frequency\_penalty}) remained at their default setting. The final dataset consists of two LLM-based corpora based on the Litkey Corpus: i) the zero-shot LLM corpus (LLM-ZS) and ii) the few-shot LLM corpus (LLM-FS) (see Table~\ref{tab:types_tokens}).

\subsection{Data Preprocessing and Comparative Analysis}
All the corpora were tokenized using the NLTK text mining library in Python \citep{bird2009natural}. We retained capitalization to preserve noun/verb distinctions in German (\textit{Lernen} vs. \textit{lernen}), maintaining the structural integrity and better comparability of the German corpora (e.g., see  \citealp{schepens2023can}). To filter out function words for the word-level analysis, we used the stopword list provided by the NLTK library \citep{bird2009natural}.

The word frequency overview included the most frequent words with their counts in each corpus and those shared between them, including a list of the 10 most frequent words with $>10$ characters (see Table \ref{tab:top_words_both}). Lexical richness was measured using the log type-token ratio (log TTR; \citealp{herdan1960type}).

\subsection{Word Frequency Estimation}
To analyze the quantitative similarity of the word frequency between the corpora, we calculated the correlation between the word frequencies based on the log-transformed word frequency. This procedure is a long-standing standard allowing optimal investigation of word frequency measures (e.g., see  \citealp{brysbaert2009moving, schepens2023can, gregorova2023access}) that reduces the dominant effect of a low number of high-frequency words (e.g., \textit{und}, and; \textit{ist}, he/she/it is; \textit{er}, he). Thus, it focuses strongly on the overall distribution of medium- and lower-frequency words. A Laplace smoothing transformation was applied to ensure all word counts were at least 1, avoiding issues with computing the logarithm of zero due to words present in one corpus but absent in the other.

\subsection{Vector-Based Semantic Analysis with Word Embeddings}
The corpus comparison on the level of semantics relies on how words are distributed in a multidimensional vector space. The analysis used vector-based semantics from word embedding models trained using neural networks (e.g., GloVe; \citealp{pennington2014glove}, Word2Vec;  \citealp{mikolov2013efficient}). Here, we use the fastText model \citep{bojanowski2017enriching} due to its ability to capture subword information, making it particularly effective for morphologically rich languages like German, smaller corpora \citep{bojanowski2017enriching}, and child language as even in the presence of spelling or orthographic errors, fastText can still capture the word's meaning \citep{grave2018learning}.
We used the preprocessed data from the Litkey Corpus and LLM-FS corpus to train a fastText model instead of using pre-trained word embeddings. After training, all words from both corpora were converted into vectors (i.e., word embeddings) by the trained fastText model. To compare the corpora, we calculated the cosine similarity for the words shared in both corpora. Computing cosine similarity within each corpus allows us to then correlate the semantic relations between words across corpora, since each corpus-specific vector space is arbitrary and cannot be compared. 

\section{Results}

\begin{table*}
  \caption{Litkey Corpus compared with zero- (LLM-ZS) and few-shot (LLM-FS) corpora based on text and token counts, lexical richness (log-TTR; token-type ratio), median word/sentence length, and word frequency correlations (indicated by the correlation coefficient r).}
  \label{tab:types_tokens}
  \centering
  \begin{tabular}{lccc}
    \toprule
    Corpus & \textbf{Litkey} & \textbf{LLM-ZS} & \textbf{LLM-FS} \\
    \midrule
    Total texts & 1,922 & 1,922 & 1,922 \\
    Total tokens & 212,505 & 363,867 & 158,785 \\
    Avg. tokens/text & 111 & 189 & 83 \\
    Total types & 6,364 & 3,855 & 2,354 \\
    log-TTR & 0.71 & 0.64 & 0.65 \\
    Median word length & 4.0 & 3.0 & 3.0 \\
    Median sentence length & 10.0 & 16.0 & 11.0 \\
    \midrule
    \textit{Word frequency correlations} & & & \\
    $r$ & & 0.47 & 0.58 \\
    $r_{\text{shared}}$ & & 0.62 & 0.66 \\
    $r_{\text{without FW}}$ & & 0.53 & 0.58 \\
    \bottomrule
  \end{tabular}
\end{table*}

\subsection{Word-Level Characteristics: Lexical Richness, Word Frequency, and Word Length}
All the corpora consist of the same number of texts (see Table~\ref{tab:types_tokens}). The LLM-ZS corpus contained the most tokens, while the LLM-FS corpus had the fewest, both in total and in average tokens per text, with the Litkey Corpus falling in between (see Table~\ref{tab:types_tokens}). In contrast, lexical richness, as indicated by the number of types and the log-TTR, was highest in the Litkey Corpus when compared to both LLM-based corpora (see Table~\ref{tab:types_tokens}). Word length was highly similar between corpora (see Figure~\ref{fig:word_sentence_length} (a) and Table~\ref{tab:types_tokens}). 

To initially compare word usage between children and LLMs, we provide the 10 most frequent words from the Litkey and LLM-FS corpora (Table~\ref{tab:long_words_llm_litkey}). Function words and protagonist names were removed. Words such as \textit{Hund} (dog), \textit{Fenster} (window), and \textit{Tasche} (bag) are frequent in both corpora. The analysis of complex words (two right columns in Table~\ref{tab:long_words_llm_litkey}) shows the most frequent words with more than ten characters. In both corpora, words such as \textit{Staubsauger} (vacuum cleaner), \textit{verschwunden} (disappeared), and \textit{erschrocken} (scared) are more specific and complex and point to more detailed descriptions. It can be seen that the model uses more general words like \textit{Bildbeschreibung} (picture description) and \textit{Klassenzimmer} (classroom). In Table~\ref{tab:top_words_both}, we go one step further and compare the 10 most frequent words shared between the two corpora. Again, the function words and names of the main characters of the picture stories were removed. Despite the shared words, there are some noticeable differences in how often these words were used. It can be noticed that words like \textit{sagt} (he/she/it says), and \textit{sagte} (said) are used much more often in the Litkey Corpus compared to the LLM-FS corpus. In contrast, words like  \textit{Hund} (a dog), \textit{Mädchen} (a girl), and \textit{sieht} (he/she/it sees) are more frequent in the LLM-FS corpus. Overall, the corpora share 22.83\% of the included types.

\begin{table*}
  \caption{Top 10 frequent words and words with more than 10 characters in the Litkey and few-shot LLM (LLM-FS) corpora (excluding function words and character names).}
  \label{tab:long_words_llm_litkey}
  \centering
  \begin{tabular}{lccc}
    \toprule
    \textbf{Litkey} & \textbf{Litkey $>$10} & \textbf{LLM-FS} & \textbf{LLM-FS $>$10} \\
    \midrule
    Hund       & Staubsauger       & Hund        & Staubsauger \\
    sagt       & verschwunden      & Mädchen     & verschwunden \\
    Fenster    & erschrocken       & sieht       & überglücklich \\
    sagte      & Staubsaugerbeutel & glücklich   & erleichtert \\
    Bus        & Fensterbank       & plötzlich   & telefoniert \\
    Mann       & weggelaufen       & Fenster     & Bildbeschreibung \\
    Eis        & Hundefutter       & Tasche      & Klassenzimmer \\
    Tasche     & Telefonnummer     & Bus         & Nachrichten \\
    Seil       & Steckbriefe       & Telefon     & erschrocken \\
    ging       & mitgebracht       & Kind        & Staubsaugerbeutel \\
    \bottomrule
  \end{tabular}
\end{table*}

\begin{table*}
  \caption{Top 10 frequent shared words in the Litkey and the few-shot LLM (LLM-FS) corpora (excluding function words and character names).}
  \label{tab:top_words_both}
  \centering
  \begin{tabular}{lccc}
    \toprule
    \textbf{Word} & \textbf{Litkey} & \textbf{LLM-FS} & \textbf{Total} \\
    \midrule
    Hund         & 975  & 2568 & 3543 \\
    Fenster      & 902  & 889  & 1791 \\
    Bus          & 762  & 833  & 1595 \\
    Tasche       & 637  & 855  & 1492 \\
    glücklich    & 424  & 1064 & 1488 \\
    sieht        & 275  & 1186 & 1461 \\
    Mädchen      & 106  & 1240 & 1346 \\
    plötzlich    & 386  & 918  & 1304 \\
    Eis          & 658  & 595  & 1253 \\
    traurig      & 459  & 693  & 1152 \\
    \bottomrule
  \end{tabular}
\end{table*}

\begin{figure}
  \centering
  \includegraphics[width=\linewidth]{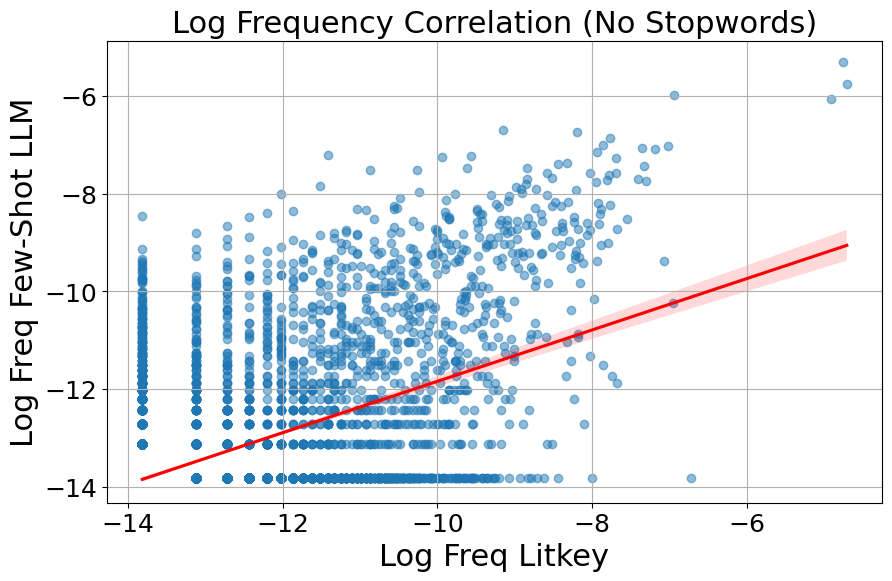}
  \caption{Log-normalized correlation between word frequencies in the few-shot LLM (LLM-FS) corpus and the Litkey Corpus, showing a correlation of $r = 0.58$ (excluding function words).}
  \label{fig:fs_litkey_correlation}
\end{figure}

Quantitative comparisons of word frequency showed that the log-frequency measured based on the Litkey is more highly correlated with the few-shot than the zero-shot corpus (see Table~\ref{tab:types_tokens}). Focusing on shared words increased similarity, as indicated by a higher correlation coefficient. Many words are absent from the LLM corpus, indicated by the low lexical richness, resulting in a lower correlation when all words are included (see Figure \ref{fig:fs_litkey_correlation}; high number of words with a log-frequency of ~0). 


\subsection{Syntax-Level Characteristics: Sentence Length and Part-of-Speech Analysis}

\begin{figure}
  \centering
  \includegraphics[width=\linewidth]{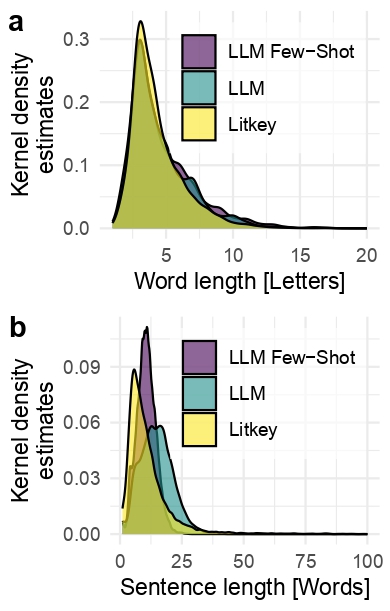}
  \caption{Distributions of (a) word length (in number of letters) and (b) sentence length (in number of words) for the Litkey, the zero-shot LLM (LLM-ZS), and the few-shot LLM (LLM-FS) corpora.}
  \label{fig:word_sentence_length}
\end{figure}

Sentence length is shortest in the Litkey Corpus and longest in the LLM-ZS corpus. The addition of two examples in the prompt of the LLM-FS resulted in an adaptation towards shorter sentence lengths, which were still longer than in the Litkey Corpus (see Figure~\ref{fig:word_sentence_length}b and Table~\ref{tab:types_tokens}). At the same time, the LLM-FS corpus has the lowest variability, as the Litkey Corpus showed shorter and longer sentences. The LLM-ZS corpus showed overall the longest sentences. Sentences over 50 words are rare but occur in the Litkey due to missing punctuation. As only spelling errors were corrected, punctuation errors remained unchanged. Twenty-nine outliers exceeded 100 words, with the longest sentence having 278 tokens, and were \mbox{excluded from the graph}.

\begin{table*}
  \caption{POS tag frequency comparison between the Litkey and the few-shot LLM (LLM-FS) corpora. Positive values indicate higher frequency in LLM-FS. NOUN: noun; VERB: verb; DET: determiner; ADV: adverb; ADP: adposition; PRON: pronoun; AUX: auxiliary; CCONJ: coordinating conjunction; SCONJ: subordinating conjunction; ADJ: adjective; PART: particle; NUM: numeral.}
  \label{tab:pos_comparison}
  \centering
  \begin{tabular}{lccc}
    \toprule
    \textbf{POS Tag} & \textbf{Litkey [\%]} & \textbf{LLM-FS [\%]} & \textbf{Diff.} \\
    \midrule
    NOUN   & 30.42 & 24.06 & -6.36 \\
    VERB   & 14.86 & 15.43 & +0.57 \\
    DET    & 10.48 & 13.34 & +2.86 \\
    ADV    & 10.87 & 12.11 & +1.24 \\
    ADP    & 8.56  & 9.59  & +1.03 \\
    PRON   & 8.48  & 8.02  & -0.46 \\
    AUX    & 6.22  & 4.78  & -1.44 \\
    CCONJ  & 5.96  & 6.22  & +0.26 \\
    SCONJ  & 0.86  & 2.82  & +1.96 \\
    ADJ    & 1.90  & 2.22  & +0.32 \\
    PART   & 0.98  & 0.88  & -0.10 \\
    NUM    & 0.41  & 0.54  & +0.13 \\
    \bottomrule
  \end{tabular}
\end{table*}

We compared POS tags between the Litkey and the lexically richer LLM-FS corpus, which was more similar to child-produced text. The Litkey Corpus contained more nouns, while all other differences were relatively small (i.e., half the size with $<$ 3\%), with the largest being more determiners and subordinating conjunctions in the LLM-generated texts (for the exact numbers, see Table~\ref{tab:pos_comparison}).

\subsection{Semantic-Level Characteristics: Vector Space Analysis}

\begin{figure}
  \centering
  \includegraphics[width=\linewidth]{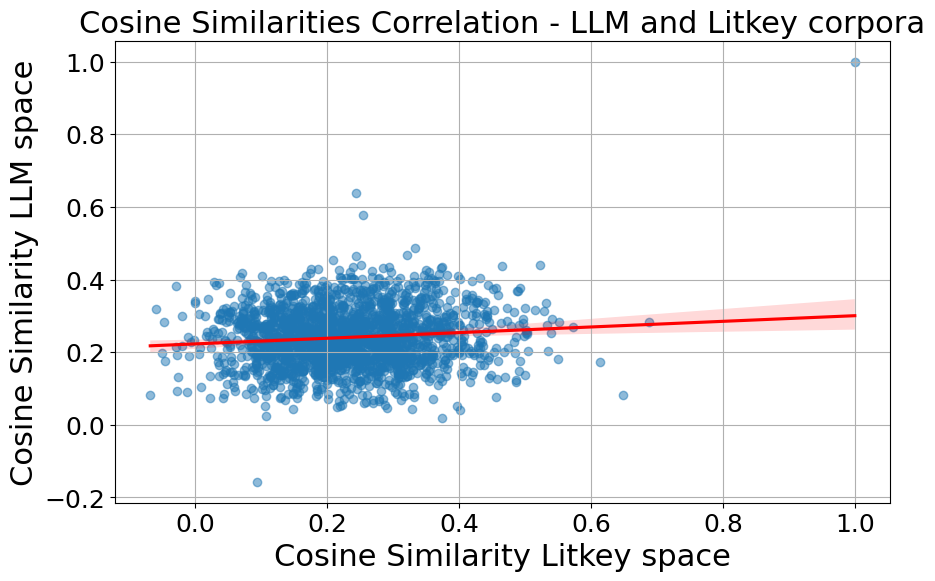}
  \caption{Cosine similarity between the Litkey and the few-shot LLM (LLM-FS) corpora based on bootstrapped vector space analysis.}
  \label{fig:bootstrapped_cosine}
\end{figure}

After training fastText-based word embeddings on the Litkey, the LLM-ZS, and the LLM-FS, we calculated the cosine similarity between all shared words within the corpora. Thus, we can compare the corpus-specific semantic spaces based on an abstract representation that is independent of the original arbitrary embeddings (see \citealp{edelman1998representation,kriegeskorte2008representational}). The correlation between the cosine similarities of the Litkey and the two LLM corpora is therefore an estimate of semantic similarity ($r$ = 0.2 for the LLM-FS, see Fig.~\ref{fig:fs_litkey_correlation}; $r$ = 0.1 for the LLM-ZS)

\section{Discussion and Limitations}
Here, we compare written text from German children describing picture stories to LLM-generated text prompted with the same images using established corpus and psycholinguistic measures. Our analysis shows that while LLMs can partially replicate children’s writing, there are significant differences on the level of words, syntax, and semantics. On the word level, LLM-based corpora involve more repetitive and less detailed vocabulary, especially nouns, resulting in lower lexical richness. The LLM-generated text contained more high-frequency and fewer mid- to low-frequency words. Correlations of word frequency showed mid to high similarity between the number of occurrences in children and LLM-generated texts. On the syntax level, LLMs produced longer sentences with fewer nouns. On the semantic level, word co-occurrence-based vector spaces differed substantially, revealing key differences in meaning representation. Thus, LLMs struggled to replicate the context-rich language of children’s storytelling despite mimicking lexical patterns and generating syntactically complex text. A few-shot prompting approach using two image-description pairs from the Litkey Corpus resulted in more similar texts regarding word frequency and sentence length, but still lacked lexical richness.

Qualitative analysis revealed that while LLMs generally captured the plot of the picture story, they still often misidentified objects and protagonists, leading to inaccuracies and hallucinations. These hallucinations, such as fabricated text on posters, may stem from the model’s tendency to process images in isolation. Frequent use of framing phrases like \textit{Auf dem Bild sehe ich...} (on the picture I can see...) contrasted with the children’s spontaneous, narrative-focused descriptions.

Lower lexical richness has previously been reported in face-to-face interactions, child-book-based corpora, and comparisons of LLMs to adult-written text \citep{liu2024benchmarking,schepens2023can,guo2023curious}. This may result from a regression-to-the-mean phenomenon, reflecting LLMs’ aim to generate broadly understandable text. \citet{schepens2023can} found that increasing temperature slightly improved lexical richness, suggesting parameter tuning as a potential step toward child language modeling. Until such parameters are established or child-specific models are developed, using LLM agents in child education is not recommended, as they may bias children’s language toward low vocabulary, potentially hindering the development of age-appropriate language skills.

Children’s books aim to enrich vocabulary \citep{dawson2021features}, and early literacy acquisition is marked by rapid vocabulary growth \citep{song2015tracing}. Educational content should use simple language only in specific contexts, such as instructional tasks, but not for creative writing or language exposure \citep{elgarf2024fostering}. Thus, we agree with \citet{liu2024benchmarking} that benchmarks need to be defined first, which should be met before LLMs are used in educational or child-directed settings.

\citet{guo2023curious} noted that LLM-generated text reduces lexical richness. One proposed solution to data scarcity is training on self-generated text \citep{wang2022self}. However, the lack of child-produced data contributes to low lexical richness and weak performance on psycholinguistic benchmarks, including parent-child interactions \citep{liu2024benchmarking}, adult-authored children’s books \citep{schepens2023can}, and children’s writing, as shown in our findings.

A key limitation of this study is the lack of transparency in the proprietary LLMs used (GPT-4V and GPT-4o). We cannot assess training data, model parameters, or architecture that potentially affect the outcomes. We have started testing open-source models, such as DeepSeek \citep{bi2024deepseek}, which are capable of modeling child language of moderate quality (i.e., see \cite{schepens2023can}), though hallucinations remain an even stronger concern. Fully open, vision-capable, and multilingual LLMs are still rare, and closed models are frequently updated without notice, hindering reproducibility \citep[e.g.,][]{maslej2024artificialintelligenceindexreport}. To resolve these issues, we need more multimodal LLMs that are open to investigation in detail and that can be frozen in a particular version. 

Another limitation is the lack of readability assessments \citep{crossley2023large}. Expanding benchmarking to include these would offer a more comprehensive evaluation. While German presents challenges due to limited child-language data, similar results in English \citep{liu2024benchmarking} suggest that these issues are not language-specific. Future work should broaden evaluation metrics and examine cross-linguistic generalizability within a single study.

Findings from our research show that, although the model can capture surface-level features of child language, its potential in educational contexts remains limited as it, in its current state, cannot generate texts in contexts involving young learners and visually grounded content. Future research may call for training on age-appropriate corpora or fine-tuning with child texts and child language tasks. Training in multiple modalities could better reflect child language, thus increasing the probability of producing child-like language, such that the models can support adequate child language development. One possibility could be to adapt the size of training data of the models to the amount possibly available at a young age, identifying potentially more realistic model architectures (i.e., as in the Baby LM challenge \citep{warstadt2025findings}), potentially including multimodal input data (e.g., see \citealp{orhan2024learning}).  

\section{Conclusions}
We find that LLM-generated descriptions of picture stories, based on prompts designed to mimic child-like language, are orthographically correct but differ psycholinguistically significantly from actual child-produced text. We found fundamental differences between children and LLM text across word (e.g., lexical richness), syntax (e.g., sentence length), and semantic levels (i.e., semantic similarity). While large language models show promise in replicating certain linguistic features of children’s language, they lack the full range of expressive and descriptive abilities. These findings highlight the need to continue exploring LLMs, triggering future developments and, more importantly, monitoring capabilities for recommendations/regulations for the usage in vulnerable groups like children. This is especially critical for closed-source models from proprietary actors in the field (e.g., see \citealp{liesenfeld2023opening} for a related discussion).
\begin{acknowledgments}
We thank Job Schepens for insightful comments on an earlier version of the manuscript.
\end{acknowledgments}

\section*{Declaration on Generative AI}
  During the preparation of this work, the author(s) used Grammarly and chatGPT for grammar and spelling check. After using these tool(s)/service(s), the author(s) reviewed and edited the content as needed and take(s) full responsibility for the publication’s content.

\bibliography{sample-ceur}

\appendix


\end{document}